\documentclass[conference]{IEEEtran}
\usepackage{cite}
\usepackage{amsmath,amssymb,amsfonts}
\usepackage{algorithmic}
\usepackage{graphicx}
\usepackage{textcomp}
\usepackage{xcolor}
\usepackage{xurl}
\usepackage{hyperref}
\usepackage{float}
\usepackage{subcaption} 
\usepackage{adjustbox}
\def\BibTeX{{\rm B\kern-.05em{\sc i\kern-.025em b}\kern-.08em
    T\kern-.1667em\lower.7ex\hbox{E}\kern-.125emX}}

\usepackage{tikz}
\usetikzlibrary{shapes.geometric, arrows.meta, positioning, calc}

\tikzstyle{startstop} = [rectangle, rounded corners, minimum width=3cm, minimum height=0.8cm, text centered, draw=black, fill=blue!5]
\tikzstyle{process} = [rectangle, minimum width=3cm, minimum height=1cm, text centered, text width=4cm, draw=black, fill=gray!5]
\tikzstyle{io} = [trapezium, trapezium left angle=70, trapezium right angle=110, minimum width=3cm, minimum height=1cm, text centered, draw=black, fill=green!5]
\tikzstyle{arrow} = [thick,->,>=stealth]

\newcommand{\authorcell}[2]{%
\begin{tabular}[t]{c}
#1\\
\textit{#2}
\end{tabular}%
}

\begin{document}

% \title{Work Zone and Temporary Speed Limit Recognition System for Real-Time Driver Assistance}

\title{\vspace{.75em}\LARGE{Vision-Language Work Zone Intelligence for Safety-Critical Speed Regulation of Mixed-Autonomy Vehicles in Dynamic Environments}}

% \author{
% \IEEEauthorblockN{Collaborator Name}
% \IEEEauthorblockA{\textit{University of California, Merced} \\
% collaborator@ucmerced.edu}
% \and
% \IEEEauthorblockN{Collaborator Name}
% \IEEEauthorblockA{\textit{University of California, Merced} \\
% collaborator@ucmerced.edu}
% \and
% \IEEEauthorblockN{Collaborator Name}
% \IEEEauthorblockA{\textit{University of California, Merced} \\
% collaborator@ucmerced.edu}
% \and
% \IEEEauthorblockN{Collaborator Name}
% \IEEEauthorblockA{\textit{University of California, Merced} \\
% collaborator@ucmerced.edu}
% \and
% \IEEEauthorblockN{Collaborator Name}
% \IEEEauthorblockA{\textit{University of California, Merced} \\
% collaborator@ucmerced.edu}
% \and
% \IEEEauthorblockN{Collaborator Name}
% \IEEEauthorblockA{\textit{University of California, Merced} \\
% collaborator@ucmerced.edu}
% \and
% \IEEEauthorblockN{Ross Greer}
% \IEEEauthorblockA{\textit{University of California, Merced} \\
% rossgreer@ucmerced.edu}
% }
\author{
% \small
\begin{adjustbox}{max width=\textwidth}
\begin{tabular}{ccccc}
\authorcell{Angel Martinez-Sanchez}{UC Merced} &
\authorcell{Kianna Ng}{UC Merced} &
\authorcell{Wesley Maia}{UC Merced} &
\authorcell{Laura Fleig}{Johns Hopkins} &
\authorcell{Maitrayee Keskar}{UC San Diego}

\\[1.5em]

\authorcell{Erika Maquiling}{UC Merced} &
\authorcell{Yash Tandon}{UC San Diego} &
\authorcell{Parthib Roy}{UC Merced} &
\authorcell{Mohan M. Trivedi}{UC San Diego} &
\authorcell{Ross Greer}{UC Merced}
\end{tabular}
\end{adjustbox}
}

\maketitle

\begin{abstract}
Temporary work-zone speed limits are communicated through visually inconsistent signage and are often missing from digital maps, creating safety risks for human drivers and automated vehicle systems. We present a real-time, onboard perception pipeline that detects active work zones, recognizes associated temporary speed limits, and outputs a law-aware work-zone state and speed value suitable for driver alerts or downstream automated control. The system fuses object detections with semantic verification and temporally smoothed, hysteresis-based state transitions to reduce false activations and flicker in dynamic scenes, and runs fully on low-cost embedded hardware. Evaluated manually on a annotated subset of the ROADWork dataset (490 sequences), the system achieves inside-work-zone event-level recall of 96.5\% and event-level precision of 68.7\%. Speed-limit recognition evaluated on 35 minutes of in-house driving data attains 95.45\% precision and 53.85\% recall, with no incorrect speed classifications and a single false positive. These results demonstrate a practical, scalable approach for grounding work-zone speed awareness directly in onboard perception rather than maps or infrastructure. We release our source code for the proposed system pipeline on our GitHub repository: \url{https://github.com/Mi3-Lab/workzone}
\end{abstract}

\begin{IEEEkeywords}
work zone safety, speed limit detection, computer vision, ADAS, autonomous vehicles, perception systems
\end{IEEEkeywords}

\section{Introduction}

Active road construction zones introduce highly dynamic driving environments that challenge both human drivers and intelligent vehicle systems. Rapidly changing roadways, temporary barriers, construction workers, and variable speed limits require continuous adaptation to evolving conditions. These environments are particularly safety-critical because vehicle speed strongly influences crash severity. For example, a pedestrian or roadway worker struck at 40~mph has only about a 20\% chance of survival, compared to approximately 90\% at 20~mph~\cite{ref_pedestrian_speed}. As a result, reliable awareness of temporary speed limits and work zone context is essential for both manual driving and automated vehicle operation.

Temporary speed limits are commonly deployed in work zones to mitigate risk under dynamic roadway conditions. These limits are often communicated through portable, electronic, or handheld displays that vary widely in appearance and are rarely documented into digital maps or navigation databases. This creates a mismatch between the physical driving environment and the internal representations used by advanced driver assistance systems and automated vehicles, resulting in incomplete situational awareness during safety-critical roadway transitions. Maintaining accurate, real-time awareness in such environments is challenging because relevant cues may change faster than conventional mapping or infrastructure-based updates can propagate.

% link to figrue: https://docs.google.com/presentation/d/1g0bSWkrV-KbqnbjkQCNmEHZAQOybf7zqN2NwHgwPDIc/edit?usp=sharing
\begin{figure}
    \centering
    \includegraphics[width=.95\linewidth]{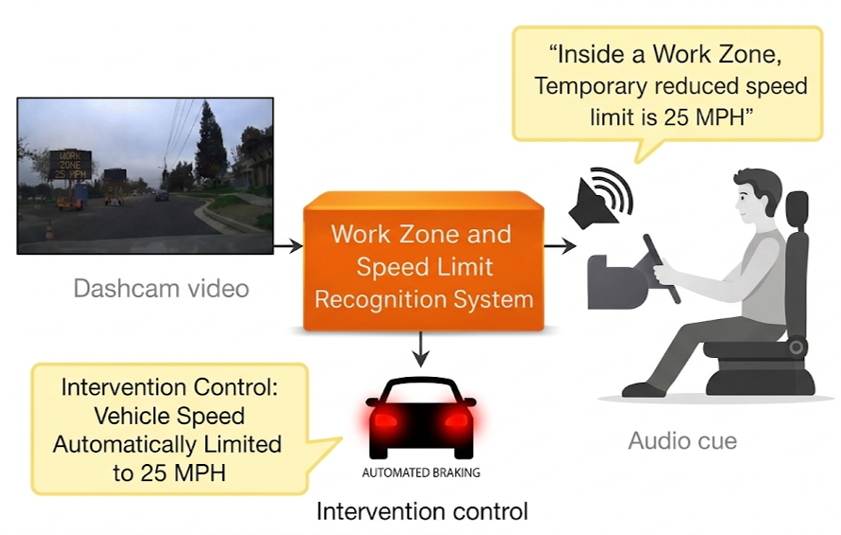}
    \caption{High-level overview of the proposed work-zone and temporary speed-limit recognition system. A live camera stream is processed onboard to detect when the vehicle is inside an active work zone and infers the applicable temporary speed limit, which can be passed downstream to AD or ADAS systems (e.g., an audio cue or intervention control).}
    \label{fig:System_Diagram}
\end{figure}

The safety consequences of these conditions are substantial. In the United States, approximately 101{,}000 crashes, 39{,}000 injuries, and 899 fatalities occurred in construction zones in 2023~\cite{ref_workzone_stats}, with speeding involved in roughly one-third of fatal incidents annually~\cite{ref_speeding_fraction}. As mixed-autonomy environments become more prevalent, vehicle control increasingly relies on machine perception rather than direct human judgment, placing greater demand on robust systems capable of recognizing dynamic roadway conditions directly from sensor input

In this work, we present a vision-based approach system that can (i) recognize active work zones, (ii) detect associated temporary speed limits directly from roadside signage, and (iii) generate real-time, law-aware outputs suitable for manual driving, mixed-autonomy supervision, and automated control as shown in Figure~\ref{fig:System_Diagram}. Rather than relying on maps or specialized infrastructure, the proposed system grounds speed awareness in direct visual observation using low-cost, off-the-shelf cameras and embedded computing already available in most vehicles. By grounding awareness in direct visual observation, we aim to reduce the mismatch between variable physical conditions and automated driving systems’ internal representations, offering a scalable pathway to improving safety in dynamic work zone environments.

% Figure~\ref{fig:final_pipeline} presents the proposed perception pipeline.  Unlike conventional sign-recognition systems that rely solely on frame-level detection, our architecture introduces parallel semantic verification and temporal state modeling. A YOLO-based detector identifies candidate cues, which are then validated using CLIP-based 
% vision-language reasoning and contextual scene classification. OCR is incorporated to extract temporary speed limits from electronic or portable signage. A multi-modal weighted fusion module applies temporal smoothing to mitigate detection flicker, and a finite-state machine enforces physically plausible transitions between work zone phases. This structured design enables robust, real-time deployment on embedded platforms.

The primary contributions of this work are as follows: (1) \textbf{Onboard work-zone and temporary speed-limit perception:} We develop a real-time perception pipeline that jointly detects active work zones and recognizes associated temporary speed limits using low-cost embedded hardware suitable for in-vehicle deployment; (2) \textbf{Temporally stable work-zone state estimation:} We introduce a temporal inference framework that fuses object detections with semantic verification and hysteresis-based state transitions to reduce flicker and false activations in dynamic scenes; (3) \textbf{Law-aware speed output interface:} We propose a unified output representation that produces actionable speed-limit with work-zone status, enabling integration with manual driving assistance, mixed-autonomy supervision, and automated control; and (4) \textbf{Open-source release:} We release our system implementation (code, weights, configurations) to support reproducibility and accelerate follow-on research.
% \end{itemize}

% \begin{itemize}
%     \item We develop a real-time perception pipeline that jointly detects active work zones and recognizes associated temporary speed limits using low-cost embedded hardware suitable for in-vehicle deployment.
    
%     \item We introduce a temporally stable work zone inference framework that combines object detection, semantic scene verification, and hysteresis-based state estimation to reduce false activations in dynamic environments.
    
%     \item We propose a unified interface for generating law-aware speed outputs applicable across manual driving, mixed-autonomy supervision, and automated vehicle control systems.
    
%     \item We provide experimental validation demonstrating reliable event-level detection performance and practical deployment feasibility on embedded computing platforms.
% \end{itemize}

\section{Related Work}

\subsection{Work Zone Behavior Across Levels of Vehicle Automation}

Prior work suggests that failures to adhere to temporary speed limits in work zones arise from different mechanisms depending on the level of vehicle automation. In manual driving, speeding behavior is often attributed to behavioral factors such as risk perception, habituation, and inconsistent enforcement rather than failures of visual perception~\cite{ref_manual_behavior}.

In partially automated vehicles, the challenge shifts toward human–automation interaction. Drivers using adaptive cruise control or lane-keeping assistance are more likely to exceed posted speed limits~\cite{ref_partial_automation_speed}, particularly when automation maintains a previously set speed and does not respond to changes in roadway context. Temporary speed limits are frequently absent from map-based speed data, and many assistance systems do not automatically adapt to temporary reductions, increasing reliance on timely driver intervention. At higher levels of automation, perception limitations become more critical. Automated driving systems depend on accurate models of roadway context, yet temporary work zone speed limits are often visually inconsistent, rapidly changing, and omitted from maps. Reports of real-world automated vehicle operation highlight cases in which temporary limits are missed or not acted upon~\cite{ref_av_missed_limits}. In such cases, failures stem not from intentional noncompliance but from incomplete situational awareness.

Advanced driver assistance system (ADAS) adoption has increased rapidly over the past decade~\cite{ref_adas_adoption}. As control authority shifts from the human driver to the vehicle, safety limitations increasingly arise from perception and system-integration constraints rather than deliberate driver behavior. These trends motivate approaches that enable vehicles to perceive temporary speed limits directly from the physical environment.

\subsection{Work Zone Detection}

Early approaches to work zone detection focused on identifying construction boundaries and roadway geometry using sensor fusion and structured environmental representations~\cite{ref_early_wz_detection}. More recently, the problem is framed as a large-scale video understanding task using deep learning~\cite{ref_dl_wz_detection}, supported by naturalistic driving datasets containing construction-related objects and scenes~\cite{ref_wz_dataset_roadwork, ref_mapillary_tsd, ref_traffic_light_survey, ref_traffic_light_detection}. Despite progress in visual detection, most existing systems formulate work zone recognition as a frame-level perception problem, focusing on detecting construction objects or determining the presence of a work zone within a scene. Prior methods generally do not model the temporal interaction between a vehicle and the work zone environment, such as distinguishing between approaching, entering, traversing, and exiting phases. As a result, perception outputs alone are often insufficient to directly support downstream vehicle behavior. In contrast, this work bridges this perception–behavior gap by modeling temporally consistent work zone awareness and leveraging it to inform law-aware driving decisions. This perspective enables visual perception outputs to be translated into actionable behavioral context, supporting safer and more compliant vehicle operation in dynamic construction environments.

\subsection{Speed Limit Detection}

Speed limit perception for permanent roadway signage has been extensively studied using both classical computer vision pipelines and deep learning–based object detectors for standardized regulatory signs~\cite{ref_speedlimit_classical, ref_speedlimit_dl}. These approaches have been incorporated into advanced driver assistance systems for speed awareness and regulation. However, work zones introduce sensing conditions that are not well addressed by methods developed for permanent regulatory signage. Temporary speed limits may be conveyed through digital speed limit (DSL) signs and temporary traffic control (TTC) message boards, which are often implemented using LED arrays. When capturing with a camera, LED flicker artifacts can arise from the interaction between LED pulse-width modulation (PWM), camera exposure, and sampling, producing temporal aliasing and spatial banding in captured frames~\cite{ref_led_flicker_deegan}. These artifacts are amplified by rolling-shutter sensors, which captures the scene sequentially across image rows causing different scanlines that can sample different phases of the LED cycle. Overall, TTC/DSL displays may appear with fragmented missing segments, truncated words, or inconsistent character visibility across frames, degrading the reliability relative to static printed signage.

Recent work on DSL signs explores multi-frame aggregation and post-processing techniques to recover numeric values~\cite{ref_dsl_multiframe}. Although the methods described proved to be accurate, they are computationally expensive as they operate at approximately 0.5~Hz. To our knowledge, we are the first to visually extract posted temporary speed limits from TTC message boards, despite their frequent use in active work zones. 

\section{Design Approach and System Requirements}

Methods to address unsafe speeds in construction zones include digital speed limit databases, infrastructure-based solutions such as smart signage, and policy or enforcement-based interventions. While effective in controlled settings, these approaches face practical limitations in temporary work zones where conditions change rapidly. Infrastructure-dependent solutions require coordination, maintenance, and investment that limit scalability and deployability across diverse roadway environments. These constraints motivate a vehicle-centric, perception-based approach that operates independently of maps or external infrastructure. The system must satisfy the following requirements:

\begin{enumerate}
    \item \textbf{Reliable, robust, and stable perception:} Accurate and temporally stable detection of work zones and temporary speed limits.
    \item \textbf{Real-time embedded operation:} Low-latency execution on embedded hardware without reliance on cloud connectivity.
    \item \textbf{Flexibility across autonomy levels:} Clear separation between perception, decision logic, and downstream actuation or driver advisories.
    \item \textbf{Law-aware behavior:} Speed limits are applied only when visibly detected, with no inference in their absence.
    \item \textbf{Safe and non-disruptive outputs:} Predictable, minimal outputs that support driver or system awareness without distraction.
\end{enumerate}

To our knowledge, no prior work integrates work zone perception, temporary speed limit recognition, and law-aware decision logic within a unified real-time pipeline designed explicitly for safety-critical deployment in dynamic construction environments.

\section{System Overview and Prototype Architecture}

\subsection{System Overview}

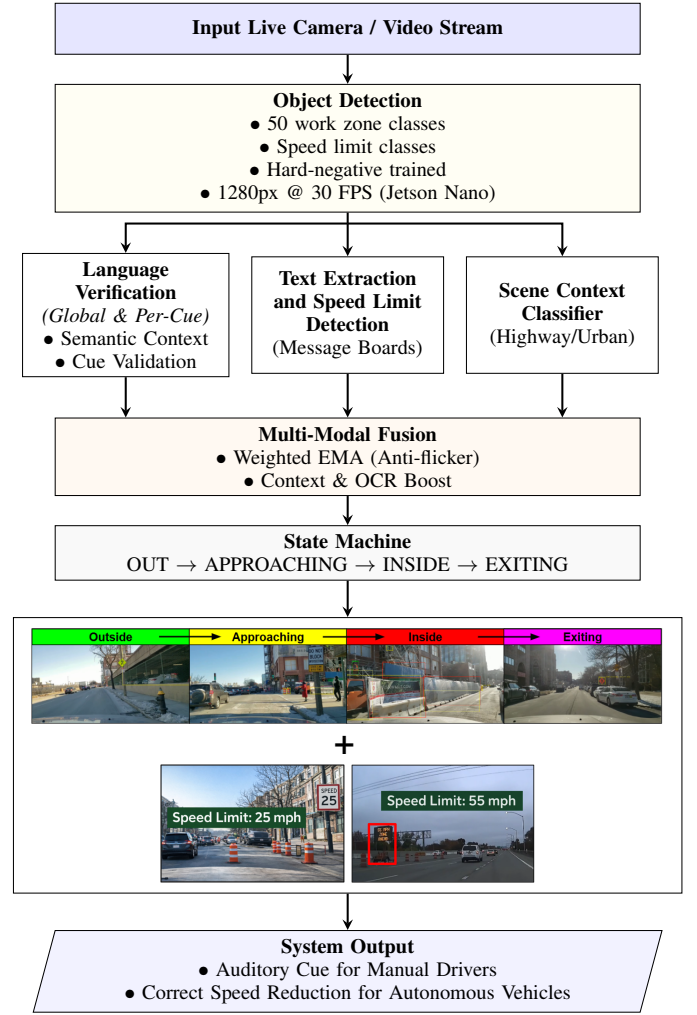
\begin{figure}
\begin{adjustbox}{max width=\linewidth}

    \centering
    \begin{tikzpicture}[node distance=0.4cm, font=\footnotesize, >=stealth]
        
        \tikzset{
            block/.style={rectangle, draw, fill=gray!5, text width=7.8cm, text centered, minimum height=0.7cm},
            parallel/.style={rectangle, draw, fill=white, text width=2.4cm, text centered, minimum height=1.6cm},
            io/.style={trapezium, draw, fill=blue!5, trapezium left angle=75, trapezium right angle=105, text width=4cm, text centered, minimum height=0.7cm},
            arrow/.style={thick, ->},
            figbox/.style={rectangle, draw, fill=white, text width=8.9cm, align=center, inner sep=4pt}
        }

        % --- Nodes ---
        \node (input) [block, fill=blue!10] {\textbf{Input Live Camera / Video Stream}};
        
        \node (yolo) [block, below=of input, fill=yellow!5] {
            \textbf{Object Detection} \\
            $\bullet$ 50 work zone classes \\
            $\bullet$ Speed limit classes \\
            $\bullet$ Hard-negative trained \\
            $\bullet$ 1280px @ 30 FPS (Jetson Nano)
        };

        \node (ocr) [parallel, below=of yolo, yshift=-0.2cm] {
            \textbf{Text Extraction and Speed Limit Detection} \\
            (Message Boards)
        };

        \node (clip_unified) [parallel, left=0.3cm of ocr, text width=2.6cm] {
            \textbf{Language Verification} \\
            \textit{(Global \& Per-Cue)} \\
            $\bullet$ Semantic Context \\
            $\bullet$ Cue Validation
        };
        
        \node (scene) [parallel, right=0.3cm of ocr] {
            \textbf{Scene Context Classifier} \\
            (Highway/Urban)
        };

        \node (fusion) [block, below=of ocr, yshift=-0.2cm, fill=orange!5] {
            \textbf{Multi-Modal Fusion} \\
            $\bullet$ Weighted EMA (Anti-flicker) \\
            $\bullet$ Context \& OCR Boost
        };
        
        % --- UPDATED: State machine box now includes + and Speed Limit Recognition ---
        \node (state) [block, below=of fusion] {
            \textbf{State Machine} \\
            OUT $\rightarrow$ APPROACHING $\rightarrow$ INSIDE $\rightarrow$ EXITING \\
        };

        % --- UPDATED: Wrap figures (+) into one boxed node, no extra spacing node-to-node ---
        \node (examplesbox) [figbox, below=of state, yshift=-0.1cm] {
            \begin{minipage}{8.7cm}
                \centering
                \includegraphics[width=8.7cm]{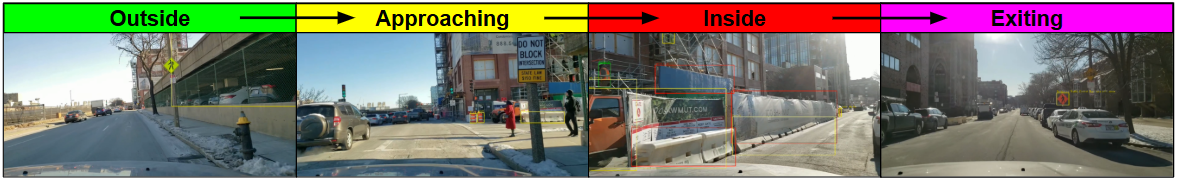}

                \vspace{0.1cm}
                {\Large \textbf{+}}
                \vspace{0.15cm}

                \includegraphics[width=0.29\linewidth]{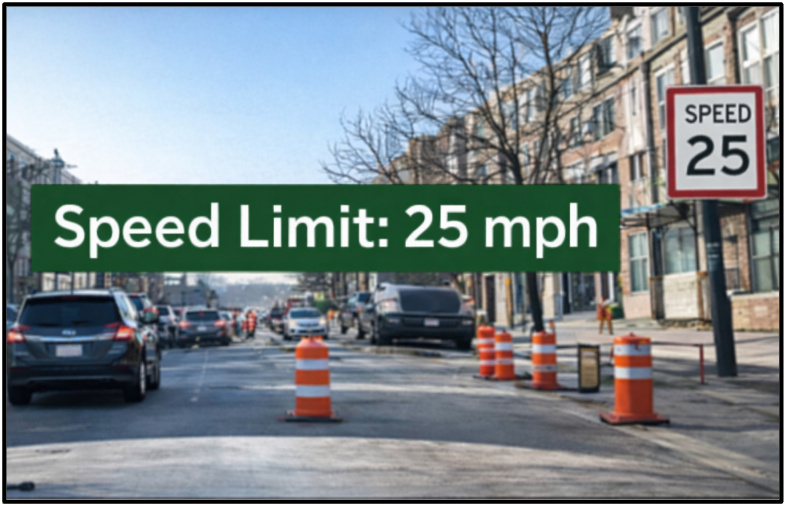} 
                \includegraphics[width=0.29\linewidth]{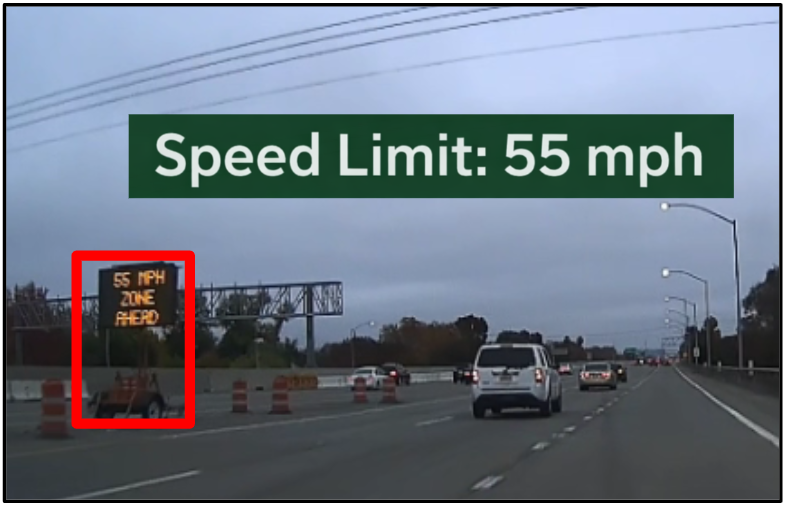}
            \end{minipage}
        };

        \node (output) [io,  text width=7.8cm,below=of examplesbox, yshift=-0.1cm] {
            \textbf{System Output} \\
            $\bullet$ Auditory Cue for Manual Drivers \\
            $\bullet$ Correct Speed Reduction for Autonomous Vehicles  \\
        };

        % --- Connections ---
        \draw [arrow] (input.south) -- (yolo.north);
        
        \draw [thick] (yolo.south) -- ++(0,-0.15) coordinate (bus);
        \draw [arrow] (bus) -| (clip_unified.north);
        \draw [arrow] (bus) -- (ocr.north);
        \draw [arrow] (bus) -| (scene.north);
        
        \draw [arrow] (clip_unified.south) -- (clip_unified.south |- fusion.north);
        \draw [arrow] (ocr.south) -- (fusion.north);
        \draw [arrow] (scene.south) -- (scene.south |- fusion.north);
        
        \draw [arrow] (fusion.south) -- (state.north);

        % route state -> boxed figures -> output
        \draw [arrow] (state.south) -- (examplesbox.north);
        \draw [arrow] (examplesbox.south) -- (output.north);

    \end{tikzpicture}
        \end{adjustbox}
    \caption{System architecture of the proposed vision-based work zone recognition pipeline. 
YOLO-based object detection feeds parallel semantic verification (CLIP), OCR-based speed limit extraction, 
and scene context classification. These modalities are fused through a temporally smoothed weighted mechanism, 
followed by a hysteresis-based state machine that produces stable, real-time work zone and speed awareness outputs.}
    \label{fig:final_pipeline}

\end{figure}

We introduce a vision-based, real-time perception pipeline that recognizes active work zones and detects associated temporary speed limits directly from the environment. The system consists of three primary stages: (1) work zone detection using temporally aggregated visual cues, (2) temporary speed limit recognition from roadside signage, and (3) generation of structured, law-aware outputs. Each stage is designed for real-time operation on embedded hardware while maintaining modular separation between perception, reasoning, and downstream interfaces. A detailed system diagram is presented in Figure~\ref{fig:final_pipeline} with its methods described in the following subsections.

\subsection{Visual Perception for Work Zone Detection}

The work zone detection module is formulated as a real-time spatio-temporal perception architecture processing a forward-facing monocular camera stream. The objective is to robustly infer the probabilistic state of an active construction environment, mitigating the high intra-class variance and visual ambiguity inherent to dynamic work zones.

\textbf{Object Detection:} The foundational layer utilizes a YOLO-based detector~\cite{ref_yolo_detector} trained on the ROADWork dataset~\cite{ref_wz_dataset_roadwork} (comprising over 5{,}000 work zone sequences and 50 target classes). For each frame at time $t$, discrete detections are semantically aggregated across $k=4$ higher-order cue groups:
\begin{equation}
    C_{\text{YOLO}}(t) = \frac{1}{|G|} \sum_{g \in G} \max_{i: c_i \in g} s_i
\end{equation}
where $s_i$ denotes the confidence of object $i$, and $G$ partitions the detection classes into mutually exclusive semantic groups (e.g., regulatory signage, channelization devices, construction vehicles, and active personnel). To systematically attenuate false positive rates induced by visually analogous artifacts, hard-negative mining is applied, iteratively retraining the detector on high-confidence background misclassifications.

\textbf{Semantic Scene Verification:} To disambiguate complex scenes, the architecture employs Contrastive Language--Image Pretraining (CLIP)~\cite{ref_clip, greer2024towards} to evaluate the global semantic coherence of the frame against a target text embedding. To optimize computational load, this verification is dynamically gated, triggering exclusively when the localized detector confidence enters a probabilistic uncertainty boundary ($C_{\text{YOLO}}(t) > \tau_{\text{CLIP}} = 0.45$):
\begin{equation}
    C_{\text{CLIP}}(t) = \cos(f_{\text{vision}}(\mathbf{x}_t), f_{\text{text}}(\text{``active construction site''}))
\end{equation}
The integrated scene-level confidence is then computed via a piecewise fusion function:
\begin{equation}
\resizebox{\columnwidth}{!}{$
C_{\text{fused}}(t) = 
\begin{cases}
C_{\text{YOLO}}(t), & \text{if } C_{\text{YOLO}}(t) \le \tau_{\text{CLIP}} \\
\alpha C_{\text{YOLO}}(t) + (1-\alpha) C_{\text{CLIP}}(t), & \text{otherwise}
\end{cases}
$}
\end{equation}
where $\alpha = 0.6$ weights the contribution of localized primitives against global semantic context.

\textbf{Temporal Smoothing and State Estimation:} The fused confidence stream is subjected to an Exponential Moving Average (EMA) filter to attenuate high-frequency noise:
\begin{equation}
    C_{\text{EMA}}(t) = \beta \cdot C_{\text{fused}}(t) + (1-\beta) \cdot C_{\text{EMA}}(t-1)
\end{equation}
with a smoothing factor $\beta = 0.3$. The filtered signal is routed into a finite state machine enforcing asymmetric hysteresis thresholds ($\tau_{\text{entry}} = 0.70$, $\tau_{\text{exit}} = 0.45$) to govern transitions across four traversal phases: $\text{OUTSIDE} \rightarrow \text{APPROACHING} \rightarrow \text{INSIDE} \rightarrow \text{EXITING}$. These parameters were derived via a constrained grid-search optimization over a validation corpus of 406 videos, formulated to maximize the event-level F1 score subject to a strict upper bound on the false positive rate ($\leq 5 \text{ events/hour}$).

% Figure~\ref{fig:system_states} illustrates representative frames corresponding to each inferred traversal phase. The qualitative examples highlight how temporally accumulated cues, rather than single-frame detections, govern state transitions. In particular, transitional ambiguity between APPROACHING and INSIDE phases is resolved through hysteresis-based persistence, preventing oscillatory behavior under intermittent visual evidence.

\begin{figure}
    \centering
    \includegraphics[width=\linewidth]{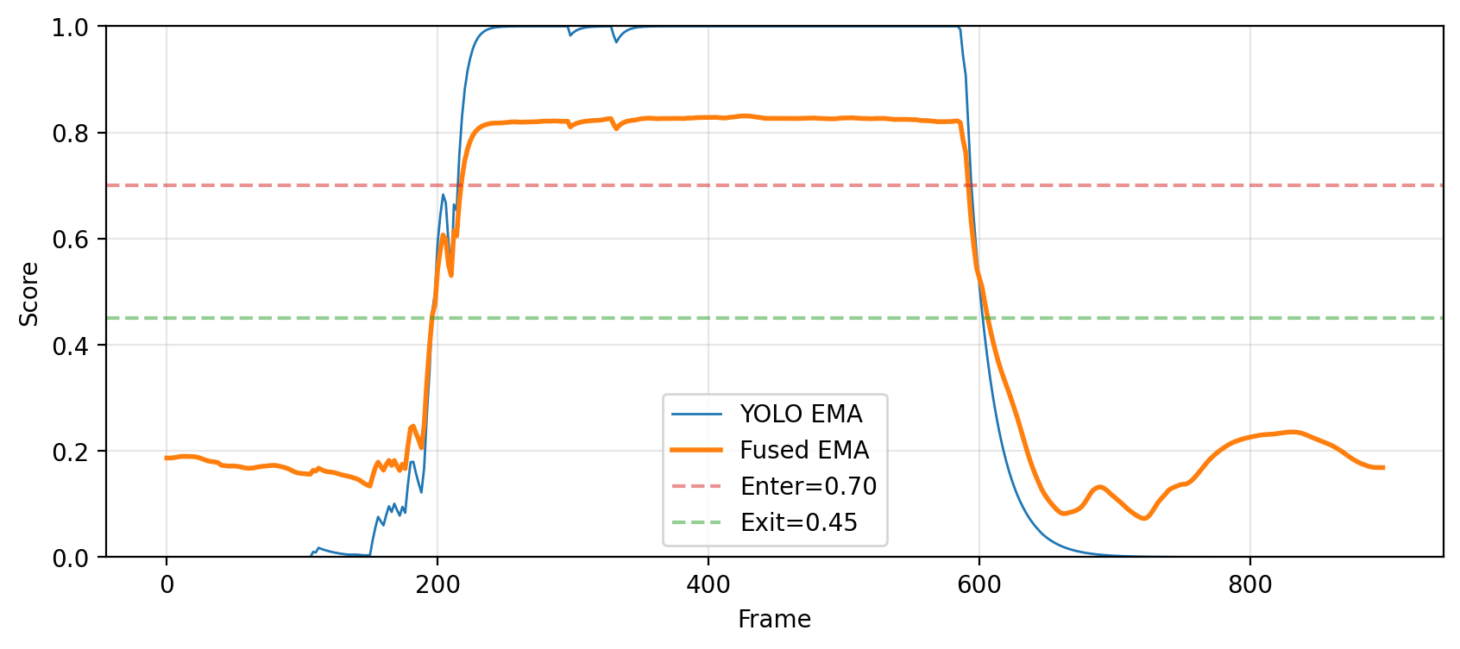}
    \caption{Temporal evolution of the work zone confidence score across a continuous drive, comparing raw YOLO object-detection outputs with the fused, EMA-smoothed signal utilized for state transitions. Dashed lines delineate the asymmetric hysteresis thresholds for state entry ($\tau_{\text{entry}} = 0.70$) and exit ($\tau_{\text{exit}} = 0.45$). The fused EMA score integrates semantic context, yielding conservative, stable values that effectively prevent state oscillation.}
    \label{fig:confidence_plot}
\end{figure}

\textbf{Preliminary Results:} Empirical evaluation demonstrates that this hybrid, temporally-smoothed architecture achieves an 84.6\% reduction in false positive detections relative to a YOLO-only baseline, yielding an F1 score of 0.87 and an operational rate of 2.6 false positives per hour on the test distribution. As shown in Figure~\ref{fig:confidence_plot}, the raw detector confidence exhibits high-frequency volatility, frequently crossing decision boundaries. In contrast, the fused and EMA-smoothed signal produces stable plateaus that align with semantically consistent roadway segments, thereby eliminating rapid threshold crossings that would otherwise induce erroneous state transitions.

% \begin{figure}[htbp]
%     \centering
%     \includegraphics[width=\linewidth]{figures/system_states.png}
%     \caption{Example real-world frames illustrating the four discrete spatial transition states—OUTSIDE, APPROACHING, INSIDE, and EXITING—as inferred from temporally accumulated visual cues. Bounding boxes denote detected construction-related object primitives. The on-screen overlay reports the inferred speed limit and current state confidence.}
%     \label{fig:system_states}
% \end{figure}

\subsection{Temporary Speed Limit Recognition}

Temporary speed limits in work zones are conveyed through heterogeneous signage types that differ in appearance, reliability, and temporal behavior. This system explicitly supports three classes of common temporary speed limit signage: (1) standard static speed limit signs, (2) digital speed limit (DSL) signs that display numeric limits using LED panels, and (3) temporary traffic control (TTC) message boards that may rotate between multiple messages.

For standard temporary speed limit signs, we employ a trained detection and classification model using publicly available traffic sign datasets, including Mapillary~\cite{ref_mapillary_tsd} and LISA traffic datasets~\cite{ref_traffic_light_survey, ref_traffic_light_detection}. The training data is augmented with examples of DSL signs, as large-scale labeled datasets for DSL signage remain limited. Hard negative mining was also used during training with background images containing similar signs and banners to reduce false detections. To ensure accurate results, a tracking window was implemented in which the sign must be continuously tracked for a minimum of 15 frames, and the detected speed limit must maintain a consistency ratio of at least 80\% in that tracking window.

For TTC message boards, the system first gets the bounding box crop from the YOLO object detector then applies optical character recognition (OCR) within the detected region. OCR outputs are aggregated across multiple frames into a history buffer that identifies keywords such as ``Speed Limit'' or ``MPH'' along with the numeric digits. These safety checks enforce plausible and consistent speed limits before the results are passed on to the downstream decision logic.

\subsection{Outputs and Human--Vehicle Interfaces}

The system's core outputs are the inferred work zone state and the perceived temporary speed limit, which may be interpreted by either human drivers or advanced driver assistance systems (ADAS). Driver-facing advisories and speed guidance are optional downstream interfaces rather than mandatory control actions. For manual and partially automated vehicles \cite{rangesh2021autonomous}, the system output can be presented as supervisory cues to support driver awareness through visual, auditory, or haptic feedback without enforcing automatic vehicle control.

For higher levels of automation, perception outputs are exposed through a structured interface for integration with planning or control modules, where detected speed limits may inform speed setpoints or motion-planning logic. Outputs may also be shared through cooperative perception frameworks to enable shared situational awareness across vehicles via vehicle-to-everything (V2X) communication. Throughout all configurations, the system maintains a clear separation between perception, decision-making, and actuation, allowing system integrators to determine appropriate enforcement strategies based on application requirements.

\subsection{Hardware and Deployment Platform}

The perception pipeline is deployed on an NVIDIA Jetson Nano paired with an Intel RealSense camera mounted inside the vehicle and runs fully onboard without cloud connectivity. In this configuration, the pipeline runs at approximately 15~FPS. The TTC message-board speed limit extraction component requires marginal additional memory and is therefore evaluated on an NVIDIA Jetson AGX Orin, where the OCR stage runs at an average of 7~FPS during active text extraction. Outside of these brief extraction intervals, the pipeline continues at its standard real-time rate. Given that the system is a software-based solution, the method is not tied to this hardware. In practice, the system can integrate with existing vehicle cameras and onboard compute resources to support broad deployment and scalability across diverse vehicle platforms
\section{Evaluation and Testing}

Evaluation focuses on the system’s ability to (1) reliably detect active work zones, (2) produce temporally stable state transitions suitable for real-time use, and (3) recognize associated temporary speed limits from roadside signage. Because missed work-zone detections are especially costly in safety-critical settings, we emphasize event-level metrics that measure whether a work zone is detected at all, alongside complementary analyses of frame-level state accuracy, transition timing error, and false-activation behavior. 

\subsection{Detection Performance}

Let $\mathcal{P}$ denote the set of predicted work zone events and $\mathcal{G}$ the set of ground-truth events. Each event is defined as a contiguous temporal interval corresponding to a specific system state (e.g., \textit{INSIDE}). A predicted event is considered a correct detection if it overlaps any ground-truth event.

We count a predicted event $p \in \mathcal{P}$ as a true positive if it overlaps any ground-truth event:
\begin{equation}
\mathrm{TP}_P =
\left| \left\{\, p \in \mathcal{P} \;\middle|\; \exists\, g \in \mathcal{G} \text{ such that } p \cap g \neq \emptyset \,\right\} \right|.
\end{equation}
Similarly, we count a ground-truth event $g \in \mathcal{G}$ as detected if it overlaps any prediction:
\begin{equation}
\mathrm{TP}_G =
\left| \left\{\, g \in \mathcal{G} \;\middle|\; \exists\, p \in \mathcal{P} \text{ such that } g \cap p \neq \emptyset \,\right\} \right|.
\end{equation}
Then,
\begin{equation}
\mathrm{Precision} = \frac{\mathrm{TP}_P}{|\mathcal{P}|},
\qquad
\mathrm{Recall} = \frac{\mathrm{TP}_G}{|\mathcal{G}|}.
\end{equation}

These event-based formulations emphasize whether work zones are detected at all, rather than frame-level accuracy, which is more appropriate for safety-critical applications where missed detections carry high cost.

Evaluation was performed on a manually annotated subset of the ROADWork dataset consisting of 490 video sequences. The system achieves an \textit{INSIDE} event recall of 96.5\% and event precision of 68.7\%, indicating that nearly all true work zone events are successfully detected while some false activations remain. Frame-level accuracy across all four states is 64.5\%, with a macro F1 score of 0.60. Per-state intersection-over-union (IoU) analysis shows strong separation between stable states (\textit{OUTSIDE} IoU 0.56, \textit{INSIDE} IoU 0.47) but substantially lower performance for boundary states (\textit{APPROACHING} IoU 0.11, \textit{EXITING} IoU 0.09), indicating that temporal transitions remain the primary challenge. 

\subsection{Temporal Behavior}

To characterize temporal alignment between predicted and ground-truth detections, we define the state entry timing offset for a given state $s$ as
\begin{equation}
\Delta t_s = t^{\text{pred}}_s - t^{\text{gt}}_s,
\end{equation}
where $t^{\text{pred}}_s$ and $t^{\text{gt}}_s$ denote the predicted and ground-truth entry times, respectively. Positive values indicate delayed detection relative to ground truth, while negative values indicate early detection. Transition performance under varying tolerance windows is summarized in Table~\ref{tab:transition_tolerance}.

\begin{table}
\centering
\vspace{.5em}
\caption{State-transition matching performance under temporal tolerance windows. A predicted transition is counted as correct if its timestamp falls within $\pm k$ frames of the corresponding ground-truth transition.}
\label{tab:transition_tolerance}
\begin{adjustbox}{max width=.95\linewidth}
\begin{tabular}{c|ccc}
\hline
\textbf{Tolerance $\pm$ (frames)} & \textbf{Precision} & \textbf{Recall} & \textbf{Accuracy} \\
\hline
0  & 0.0215 & 0.1427 & 0.0133 \\
5  & 0.0512 & 0.1931 & 0.0429 \\
15 & 0.1136 & 0.3056 & 0.1038 \\
30 & 0.1918 & 0.4272 & 0.1815 \\
\hline
\end{tabular}
\end{adjustbox}
\end{table}

Across evaluated sequences, advisory activation timing exhibits a mean absolute error of 81 frames (2.7~s), with a median error of 39 frames (1.3~s), indicating that most detections occur within a short temporal window while a small number of large deviations increase the mean. Entry timing for the \textit{INSIDE} state shows a mean absolute error of 142 frames (4.7~s), reflecting variability in boundary localization.

Because transition boundaries are inherently ambiguous and subject to annotation variability, transition detection was evaluated under temporal tolerance windows of $\pm\{0,5,15,30\}$ frames. Transition recall and precision increase with tolerance as expected, reaching 0.43 recall and 0.19 precision at $\pm30$ frames. This trend indicates that predicted transitions are generally localized near the correct temporal region even when exact frame alignment differs. The remaining precision gap suggests prediction fragmentation near state boundaries, consistent with the reduced IoU observed for \textit{APPROACHING} and \textit{EXITING} states.

\subsection{Calibration and Parameter Selection}

System parameters are calibrated using a validation dataset of 406 diverse work zone videos spanning highway construction, urban lane closures, and shoulder work scenarios. Calibration is formulated as a constrained optimization problem: maximize event-level F1 score while constraining false positive rate to $\leq$5 events per hour. Detection stability is additionally quantified using a flicker rate metric, defined as state transitions per minute, and configurations with elevated flicker are penalized to improve usability. 
% \begin{equation}
% \text{flicker\_rate} = \frac{\text{state transitions}}{\text{duration (minutes)}}
% \end{equation}

CLIP verification is selectively triggered when aggregated YOLO confidence exceeds 0.45, corresponding to an uncertainty region where object-level evidence suggests a work zone but scene-level context improves precision. Hysteresis thresholds are asymmetric (entry: 0.7, exit: 0.45) to suppress flicker and preserve state persistence through brief occlusions.

\subsection{Ablation Study}

Table~\ref{tab:ablation} summarizes the contribution of major system components. Compared to a YOLO-only baseline, CLIP verification and temporal hysteresis substantially reduce false activations, and the full system configuration achieves the best event-level performance under the false-positive constraint.

\begin{table}

\centering
\vspace{.5em}
\caption{Ablation study evaluating component contributions to event-level performance.}
\label{tab:ablation}
\begin{adjustbox}{width=.5\textwidth}
\begin{tabular}{lccl} 
\hline
\textbf{Configuration} & \textbf{Event F1} & \textbf{FP Rate (hr$^{-1}$)} & \textbf{Notes} \\
\hline
YOLO only (baseline) & 76.9\% & 12.3 & High false positive rate \\
+ CLIP verification & 85.6\% & 7.8 & Major FP reduction \\
+ Hysteresis (0.7/0.45) & 89.4\% & 5.1 & Flicker suppression \\
Full system (+ EMA) & 91.2\% & 4.2 & Production configuration \\
\hline
\end{tabular}
\end{adjustbox}
\end{table}
\subsection{Speed Limit Recognition Performance}

The speed limit recognition performance was evaluated using approximately 35 minutes of in-house driving data, encompassing a total of 39 posted 25 mph and 55 mph static signs within construction zones. As we prioritized  reliable correct detections, a precision of 95.45\% with only one false positive and zero incorrect speed classifications were achieved within our in-house dataset. Our tracking window method also resulted in a recall of 53.85\% due to unstable predictions during poor visibility conditions such as nighttime rain or heavy fog. Despite the missed detections, the system successfully recognized at least one speed limit sign in all but one of the 10 evaluated driving clips within the first or second appearance of a posted speed limit sign.

\begin{figure}
\begin{adjustbox}{max width=\linewidth}

  \centering

  \begin{subfigure}[b]{0.32\columnwidth}
    \centering
    \includegraphics[height=3.5cm,keepaspectratio]{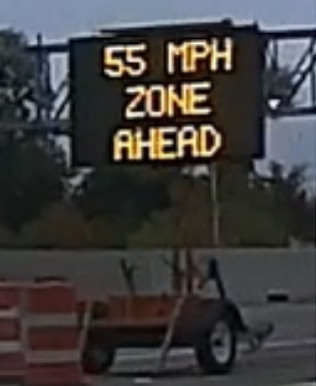}
    \caption{Clear Capture}
    \label{fig:ttcmb_captures_clear}
  \end{subfigure}\hfill
  \begin{subfigure}[b]{0.32\columnwidth}
    \centering
    \includegraphics[height=3.5cm,keepaspectratio]{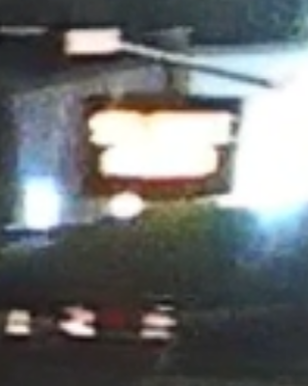}
    \caption{Blurry Capture}
    \label{fig:ttcmb_captures_blurry}
  \end{subfigure}\hfill
  \begin{subfigure}[b]{0.32\columnwidth}
    \centering
    \includegraphics[height=3.5cm,keepaspectratio]{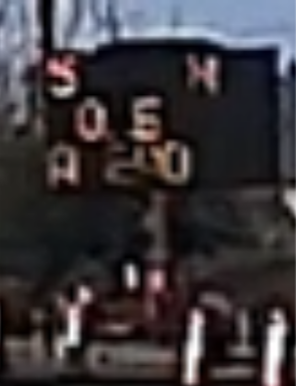}
    \caption{Partial Capture}
    \label{fig:ttcmb_captures_partial}
  \end{subfigure}
\end{adjustbox}
  \caption{Examples of TTC message-board captures under varying visibility and display artifacts.}
  \label{fig:ttcmb_captures}
\end{figure}

While the captured driving clips did not contain DSL signs, we did encounter TTC message boards. Across seven TTC message board instances, three were clearly captured and four exhibited degraded legibility (two blurry and two partial). When the camera captured a clear, complete message, the system achieved a 100\% success rate in extracting the correct speed limit. However, the current pipeline could not reliably recover text from blurry or partially rendered messages, as illustrated in Figure~\ref{fig:ttcmb_captures}. Image blurring primarily resulted from adverse weather such as rain and low-light conditions compounded by the camera's placement inside the cabin behind the windshield, which introduced glare and reflections. Furthermore, partial text captures occurred frequently due to asynchronous timing between the TTC board LED pulse-width modulation and the camera's rolling shutter effect.

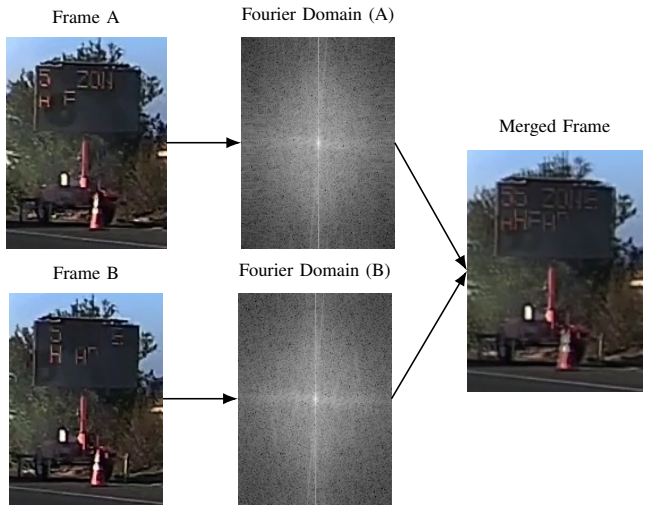
\begin{figure}
  \centering
  \begin{adjustbox}{max width=.95\linewidth}

  \begin{tikzpicture}[
    >=Latex,
    img/.style={inner sep=0pt, outer sep=0pt},
    arr/.style={-Latex, line width=0.6pt},
    lab/.style={font=\scriptsize, inner sep=1pt, outer sep=0pt}
  ]

  \def\imgh{2.8cm}
  \def\xsep{10.0mm}
  \def\ysep{6.0mm}

  \node[img] (in1)  {\includegraphics[height=\imgh,keepaspectratio]{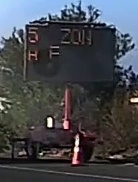}};
  \node[img, below=\ysep of in1] (in2)
                 {\includegraphics[height=\imgh,keepaspectratio]{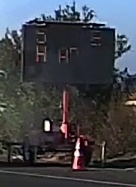}};

  \node[img, right=\xsep of in1] (mid1)
                 {\includegraphics[height=\imgh,keepaspectratio]{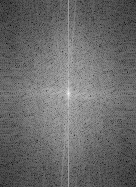}};
  \node[img, right=\xsep of in2] (mid2)
                 {\includegraphics[height=\imgh,keepaspectratio]{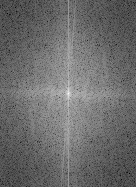}};

  \node[img, right=2.0 of $(mid1)!0.5!(mid2)$] (out)
                 {\includegraphics[height=3.2cm,keepaspectratio]{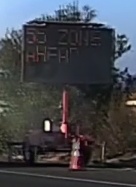}};

  % ---- Labels above each image ----
  \node[lab, above=1.5mm of in1]  {Frame A};
  \node[lab, above=1.5mm of in2]  {Frame B};

  \node[lab, above=1.5mm of mid1] {Fourier Domain (A)};
  \node[lab, above=1.5mm of mid2] {Fourier Domain (B)};

  \node[lab, above=1.5mm of out]  {Merged Frame};

  % ---- Arrows ----
  \draw[arr] (in1.east) -- (mid1.west);
  \draw[arr] (in2.east) -- (mid2.west);
  \draw[arr] (mid1.east) -- (out.west);
  \draw[arr] (mid2.east) -- (out.west);

  \end{tikzpicture}
  \end{adjustbox}
  \caption{Multi-frame reconstruction for partially captured TTC message boards. Consecutive crops are aligned via Fourier-domain phase correlation and combined using a per-pixel median merge to improve human readability.}
  \label{fig:pwm_merge}
\end{figure}

To mitigate partial LED captures on TTC message boards, a lightweight multi-frame reconstruction method was implemented. For each detection, a sliding window of consecutive crops (e.g., $N=2$ or $N=3$) is maintained. Neighboring frames are aligned to a reference crop using Fourier-domain phase correlation and combined via a per-pixel median merge as shown in Figure~\ref{fig:pwm_merge}. Although this computationally inexpensive approach produces a visually improved human-readable composite, the merged outputs did not consistently produce OCR readings with sufficient confidence for automated speed-limit extraction. In particular, consecutive frames often lacked the complementary LED strokes needed for full text reconstruction. Attempting to use temporally distant frames to fill these gaps also proved ineffective as continuous motion of the vehicle introduced significant perspective skew that our translation-only alignment could not resolve, resulting in fragmented OCR outputs.

\section{Limitations, Future Directions, and Concluding Remarks}

This work demonstrates the feasibility of perception-based recognition of work zones and temporary speed limits using forward-facing visual perception and embedded computing.

Limitations which motivate future research include resolving inherent ambiguity of work-zone boundaries and transitions, and degradation of transition timing  under occlusion, clutter, and annotation variability. Also, temporary speed-limit signage remains difficult to confidently classify, especially while TTC message boards are sensitive to blur, glare, low-light conditions, and rolling-shutter/PWM artifacts that can yield partial text captures and reduce OCR confidence. These limitations can be overcome in future research, including expansion of training and evaluation data to better cover adverse weather, and nighttime conditions. Incorporation of stronger temporal reasoning may improve boundary localization and reduce fragmentation near state transitions. Additional research directions include improved multi-frame reconstruction and display-aware processing for LED message boards. Finally, cooperative perception mechanisms such as vehicle-to-everything (V2X) communication could enable shared work-zone awareness \cite{greer2023pedestrian, denk2022assessment} when signage is partially occluded or not directly visible.

Overall, the proposed system illustrates a practical and scalable pathway for improving work-zone safety for intelligent transportation systems. By grounding speed awareness in direct visual perception rather than static digital representations, the approach better aligns vehicle behavior with rapidly changing roadway conditions. Its low-cost, hardware-agnostic design supports deployment across manual, mixed-autonomy, and automated vehicles, offering a near-term opportunity to reduce speeding and increase driver alertness in dynamic construction environments.

\bibliographystyle{IEEEtran}
\bibliography{refs}

\end{document}